# Improvements of the 3D images captured with Time-of-Flight cameras.

*D. Falie*
Politehnica University of Bucharest, Romania.
dfalie@alpha.imag.pub.ro,

**Abstract:** 3D Time-of-Flight camera's images are affected by errors due to the diffuse (indirect) light and to the flare light. The presented method improves the 3D image reducing the distance's errors to dark surface objects. This is achieved by placing one or two contrast tags in the scene at different distances from the ToF camera. The white and black parts of the tags are situated at the same distance to the camera but the distances measured by the camera are different. This difference is used to compute a correction vector. The distance to black surfaces is corrected by subtracting this vector from the captured vector image.

## Introduction.

The Time of Flight (ToF) cameras are 3D movie cameras manufactured by a few companies: MESA Imaging, PMD, Canesta and 3DV Systems. The presented method was tested on the ToF camera SR3100 manufactured by MESA.

Each pixel of the ToF camera measures the distance to the object whose image was focused on it. The distance information is obtained by illuminating the scene with a high frequency amplitude modulated infrared light (IR). In front of the sensor chip is placed a narrow band IR filter which greatly attenuates the light generated by other sources, and camera's pixels mainly detects the amplitude modulated light emitted by it and reflected by the objects in the scene. The amplitude of the detected light gives a usual black and white image and the phase gives the distance image [1], [2].

A camera's pixel synchronously detects the incoming light signal and four consecutive samples are acquired ($S_1$, $S_2$, $S_3$ and $S_4$). From these four samples are computed the two main components of the detected signal $I_y(r,c) = S_1 - S_3$ and $I_x(r,c) = S_2 - S_4$, which are the projections of a vector ($I(r,c)$) on *x* and *y* axis. The amplitude $a(l,c)$ and phase *φ(r,c)* of this vector detected by a pixel situated on raw *r* and column *c* is:

$$a(r,c) = \sqrt{I_x(r,c)^2 + I_y(r,c)^2} \qquad (1)$$

$$\varphi(r,c) = arc\tan \frac{I_y(r,c)}{I_x(r,c)} \qquad (2)$$

The phase *φ(r,c)* is the phase difference between the modulation envelope of the emitted and received light signals. Knowing φ(r,c) the distance to the object from which the light was reflected *d(r,c)* is computed with (3) where $c_0$ is the speed of light and f and is the modulation frequency.

$$d(r,c) = c_0 \cdot \frac{\varphi(r,c) + \pi}{4 \cdot \pi \cdot f} \qquad (3)$$

The measured distance can be wrong if a part of the light reflected by an object does not come directly from the illuminating source and is reflected by other objects in the scene. This indirect or diffuse light has travelled a longer path and the measured distance is greater than the real one [3], [4], [5]. Furthermore, inside the camera body the incoming light isn't totally absorbed by the sensor's pixels and some light is reflected, after multiple reflections a fraction of this light is detected by other pixels and produces a similar error. In the case of scattered light inside the camera body (flare light) the measured distance can be smaller or greater than the actual one depending on the travelled distance of the flare light. In the case of normal cameras this phenomenon produces artefacts called flare.

The errors produced by the diffuse and flare light affects differently the objects, in the distance image. There are cases when the measured distance to fixed object changes when other objects moves and this can happen even in the case when these object moves outside the image's frame. These particular situations point out that the distance image cannot be corrected having only the captured images and is necessary to have additional information. The presented correction method for the distance images uses additional information about the tags placed in the scene such as: the white and black parts of the tag are situated at the same distance to the camera or knowing the reflectivity ratio between the white and black parts of the tag. Obviously, this additional information is not enough for perfectly correct the whole distance image and adding many tags the correction is improved.

## 1. The correction method using high a contrast tag.

This method corrects mainly the errors produced by the flare light and some errors produced by the diffuse light on glossy surfaces. The flare light affects more the dark black objects than the white objects and the measured distance to an object depends on its reflectivity. The diffuse light can produce a similar error if the tag's surface is glossy and black glossy objects are affected more than white objects. These errors are caused by a perturbing signal produced by the flare and diffuse light and the problem is to find the value of the signal which caused the measured distance error between the white and black part of the tag.

There an infinite number of perturbing vectors, which can cause the same measured distance error. If in the scene is placed a mate tag with a calibrated reflectivity is possible to compute the value of the perturbing vector. Another more practical option is to use the perturbing vector with the minimum amplitude value which corrects the distance error between the white and black parts of the tag. The 3D image in the neighbourhood regions of the tag is reasonably corrected by subtracting this vector from the vector image.

The amplitude and distance images captured with the ToF camera are converted to a vector image using (3). This vector image has the components $I_x(r,c)$ and $I_y(r,c)$ on axis $x$ and $y$. The value of these components to the white and black zones of the tag are: $I_{xw}$, $I_{yw}$, $I_{xb}$, $I_{yb}$. The components of the correction vector with the minimum amplitude value are:

$$I_{cx} = -\frac{\dfrac{I_{xb} \cdot I_{yw} - I_{xw} \cdot I_{yb}}{I_{xb} - I_{xw}}}{\dfrac{I_{yb} - I_{yw}}{I_{xb} - I_{xw}} + \dfrac{I_{xb} - I_{xw}}{I_{yb} - I_{yw}}}, \qquad I_{cy} = \frac{\dfrac{I_{xb} \cdot I_{yw} - I_{xw} \cdot I_{yb}}{I_{yb} - I_{yw}}}{\dfrac{I_{yb} - I_{yw}}{I_{xb} - I_{xw}} + \dfrac{I_{xb} - I_{xw}}{I_{yb} - I_{yw}}} \qquad (4)$$

The components of the corrected vector image $I'_x(r,c)$ and $I'_y(r,c)$ are computed with (5),(6), and the amplitude and distance images with (1), (2) and (3).

$$I'_x(r,c) = I_x(r,c) - I_{cx} \qquad (5)$$

$$I'_y(r,c) = I_y(r,c) - I_{cy} \qquad (6)$$

From (4) can be observed that the correction vector cannot be computed if the one the denominators is zero and this is the case when the measured distance to the white region of the tag is equals to that to the black region. This case was expected because can not be corrected an un-distorted 3D image. When $I_{xw} = I_{xb}$ but $I_{yw} \neq I_{yb}$ then $I_{cx} = I_{xw} = I_{xb}$ and $I_{cy} = 0$, and when $I_{yw} = I_{yb}$ but $I_{xw} \neq I_{xb}$ then $I_{cy} = I_{yw} = I_{yb}$ and $I_{cx} = 0$.

## 2. The correction method using two high contrast tags.

Improvement of larger regions of the image is possible using more tags, but in this situation is necessary to interpolate the two corrections. In the situation when are used two tags can be computed a correction vector which (perfectly) corrects the image of the two tags. This correction vector is computed using the distance and amplitude images of two half white half black tags placed at different distances in the scene. The corrected image is obtained by subtracting the perturbing vector from the vector image (5),(6). Because the flare and diffuse light have a slow spatial variation the image is also improved in the neighbourhood regions of these tags.

The pixel's value is averaged in two close small regions of the white and black parts of the tags. For the first and the second tag, these values are: $I_{xw1}$, $I_{yw1}$, $I_{xb1}$, $I_{yb1}$ and $I_{xw2}$, $I_{yw2}$, $I_{xb2}$ $I_{yb2}$. Using these values the components of the correction vector $I_{cx}$ and $I_{cy}$ are computed.

$$I_{cx} = -\frac{(I_{xb1} \cdot I_{yw1} - I_{xw1} \cdot I_{yb1}) \cdot (I_{xb2} - I_{xw2}) - (I_{xb2} \cdot I_{yw2} - I_{xw2} \cdot I_{yb2}) \cdot (I_{xb1} - I_{xw1})}{(I_{yb1} - I_{yw1}) \cdot (I_{xb2} - I_{xw2}) - (I_{xb1} - I_{xw1}) \cdot (I_{yb2} - I_{yw2})} \quad (7)$$

$$I_{cy} = -\frac{(I_{xb1} \cdot I_{yw1} - I_{xw1} \cdot I_{yb1}) \cdot (I_{yb2} - I_{yw2}) - (I_{xb2} \cdot I_{yw2} - I_{xw2} \cdot I_{yb2}) \cdot (I_{yb1} - I_{yw1})}{(I_{yb1} - I_{yw1}) \cdot (I_{xb2} - I_{xw2}) - (I_{yb2} - I_{yw2}) \cdot (I_{xb1} - I_{xw1})} \quad (8)$$

The $I'_x(r,c)$ and $I'_y(r,c)$ components of the corrected vector image are computed with (5),(6), and the amplitude and distance images with (1), (2) and (3).

The components $I_{cx}$ and $I_{cy}$ can be computed with (7) and (8) if the denominators are different from zero, and these conditions are satisfied if the tags are situated at different distances from the camera, and the measured distance to the white and black parts of the tags are different (the tag's 3D image is distorted). The correction can be extended to the neighbouring regions if the amplitude of the correction vector is smaller than the average amplitude these regions, other ways the resulting image may be more distorted than the initial one, because, generally, the average amplitude of the perturbing vector caused by the diffuse and flare lights is always much smaller than the average amplitude of the directly reflected light. The computed correction vector must satisfy this condition:

$$\sqrt{I_{cx}^2 + I_{cy}^2} \ll average[I(r,c)] \quad (9)$$

## 2. Experimental Results.

The scene used to show how this method work is represented in Fig. 1. The background is a wall on which are placed different tags with different reflectivity. In front of the bear's left leg is placed also a black and white tag. The 3D image of the black objects is visibly distorted. The distance images captured with the ToF camera show some noise in the region of black objects and for this demonstration the amplitude and the distance images were filtered.

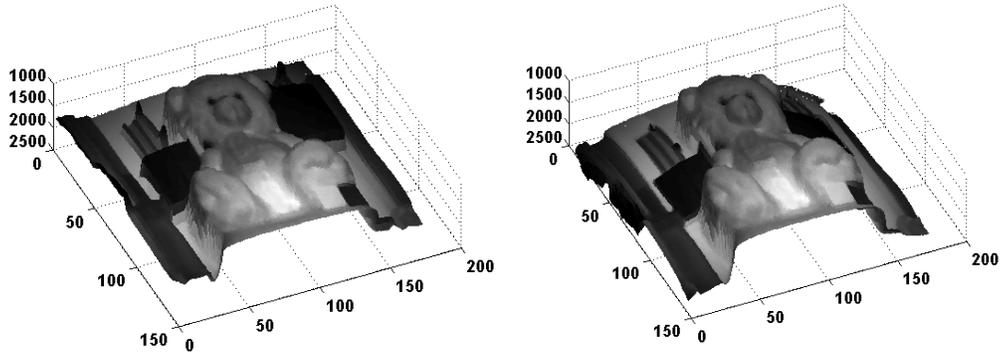
Figure 1, the 3D image of the scene (left) and the improved image (right).

The 3D image is improved in the region of the tag highlighted in Fig.2-left. The distance measured to the white part of the tag is 1760mm and 1400 mm to the black part. In Fig. 2-right is represented the distance image of the scene, in this image the black objects are closer to the camera and the gray scale at the right can be used to evaluate the image distortions of black objects.

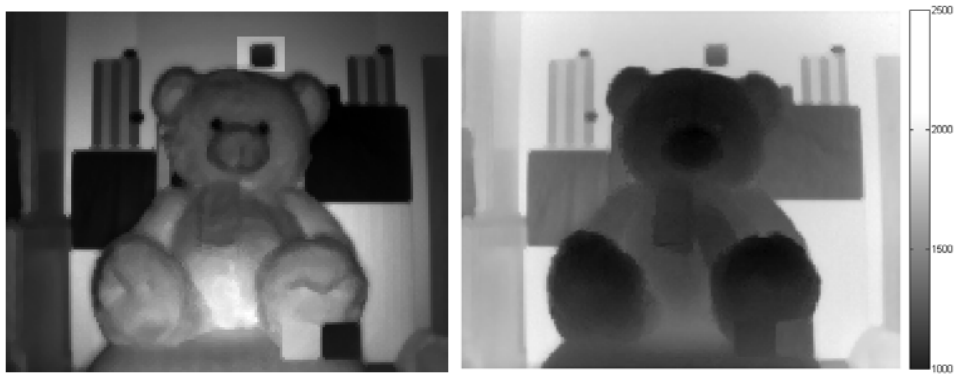
Figure 2, the amplitude image of the scene (left) and the distance image (right).

The corrected 3D image in Fig. 1-right shows a visible improvement in the region of black objects.

In the distance image represented in Fig. 3-left are marked the two high contrast zones used to compute the correction vector. The black surfaces on the background wall appear closer than the rest of the wall's surface, but the image of tag, placed in front of the left leg, is distorted in an opposite direction, the white part is closer and appears black in the distance image, Fig. 2-right.
The selection of the high contrast zones can be done preferably using the distance image because on this image are visible the distance errors. It is necessary that the measured distance to the white and black regions of the tag tot be different otherwise the denominator of some fractions in (7) and/or (8) is zero and the correcting vector cannot be computed.

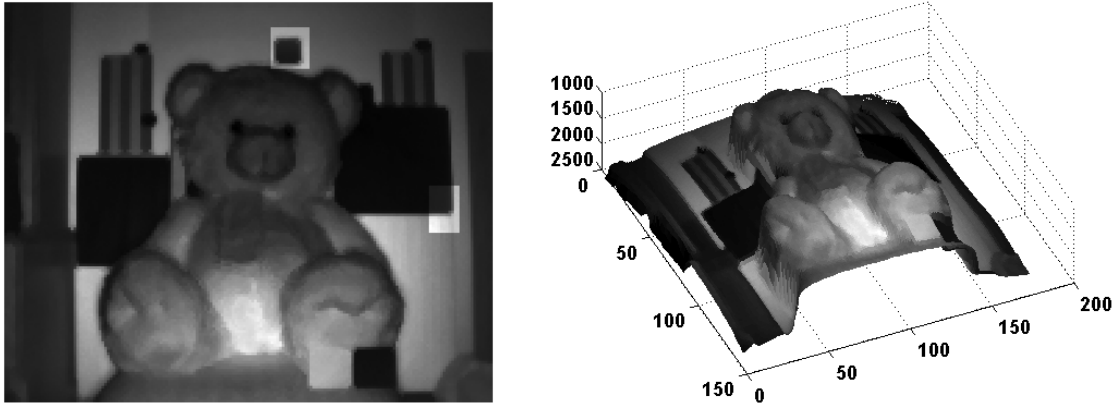

Figure 3, the selected zones are highlighted with white rectangles (left), and (right) the corrected 3D image.

The 3D image of the tags placed on the background wall was visibly improved in the corrected 3D image represented in Fig. 3-right. As It was expected the whole image cannot be corrected because the diffuse and flare lights have a slaw spatial variation, but these are not constant. The black part of the tag placed in front of the left leg is more distorted than in the uncorrected image and also the bear's eyes.

Successive corrections can be applied to the 3D image but better results are obtained correcting separately different image planes. The 3D image of the background objects in Fig. 3-right was reasonably improved but this correction distorts the bear's image. This image can be used to separate (by segmentation) the bear's image from the background because here the bear is closer than 1500mm and all the other objects are more distant. In the initial image, some black labels placed on the wall appear closer than the bear and is much more difficult to use this image for segmentation.

In the new corrected image in Fig 4-right the background image is corrected using the high contrast zones selected in Fig. 3-left, and the bear's image is corrected using the selected zone highlighted in Fig. 4-left. The bear's eyes are black and glossy and the 3D they appear inside the face plane, the corrected 3D image in Fig. 4-right shows a significant improvement.

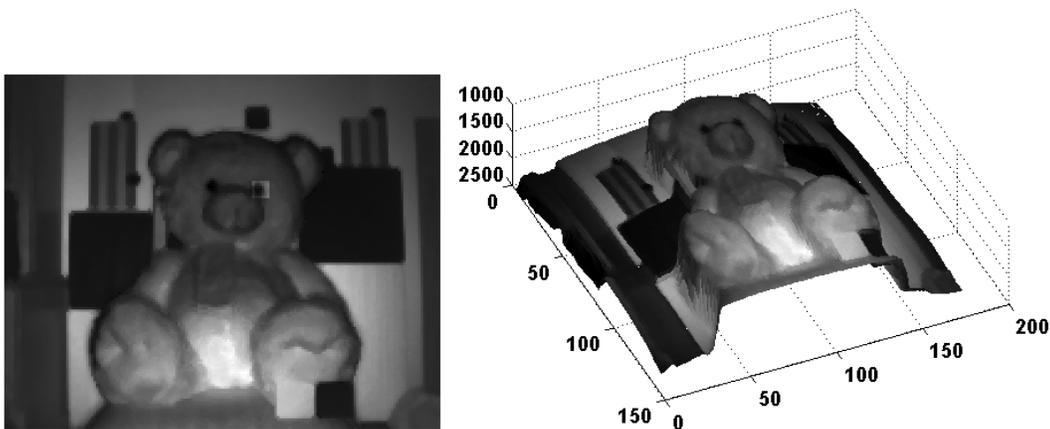

Figure 4, the selected zone is highlighted in the left image, the corrected 3D image (right).

The 3D image can be further improved by another corrected using the tag placed in front of the left leg and other distance segmentation. The MATLAB code and the test image are available at the following address: http://alpha.imag.pub.ro/~dfalie/software.html.

## Conclusions.

These methods for correcting the 3D ToF camera images give good results, and it will be very helpful in industrial and other applications where higher distance accuracy is required [13].

With this method, the camera calibration can easily be performed in any laboratory [7, 8, 9, 10, 11, and 12]. The camera calibration in "factory" is not an easy task even is realized in severe conditions where the all the room's walls are painted in a dark back to minimize the diffuse light [8, 9, and 10].

Instead of the high contrast tags can be used two or more spot lights modulated with the same frequency as the original illuminating system. In one frame a spot light is on and in the consecutive frame the other. The additional information obtained can help to improve the whole image.

## References.

[1] Z. Xu, R. Schwarte, H. Heinol, B. Buxbaum, T. Ringbeck, Smart pixel – photonic mixer device (PMD), in: Proc. Int. Conf. on Mechatron. & Machine Vision, 259–264, 1998.W.-K. Chen, *Linear Networks and Systems* (Book style). Belmont, CA: Wadsworth, 1993, pp. 123–135.

[2] R. Lange, 3D Time-Of-Flight Distance Measurement with Custom Solid-State Image Sensors in CMOS/CCD-Technology, Ph.D. thesis, University of Siegen, 2000.

[3] MESA Imaging SA, SwissRanger SR-3000 Manual Version 1.02 October, 2006.

[4] S. Oprisescu, D. Falie, Mihai Ciuc, Vasile Buzuloiu, "Measurements with ToF Cameras and their necessary corrections" ISSCS, Iassy, Romania, 2007.

[5] D. Falie, V. Buzuloiu, "Wide Range Time of Flight Camera for outdoor surveillance", 2008 Microwaves, Radar and Remote Sensing Symposium (MRRS-2008), September 22-24, 2008 Kiev, Ukraine (pp. 79-82).

[6] D. Falie, V. Buzuloiu, "Distance Errors Correction for the Time of Flight (ToF) Cameras", IEEE Workshop on Imaging Systems and Techniques (IST2008), (pp. 123-126), Chania, Crete, Greece, 2008.

[7] D. Falie,"3D Image Correction for Time of Flight (ToF) Cameras", 7156-132, Int. Conf. of Optical Instrument and Technology- OIT08, Beijing – China, 2008.

[8] M. Lindner, A. Kolb, Lateral and Depth Calibration of PMD-Distance Sensors, in: Int. Symp. on Visual Computing (ISVC), vol. 2, Springer, LNCS, 524–533, 2006.

[9] T. Kahlmann, F. Remondino, H. Ingensand, Calibration for Increased Accuracy of the Range Imaging Camera SwissRangerTM, in: Image Engineering and Vision Metrology (IEVM), 2006.

[10] M. Lindner, A. Kolb, Calibration of the intensity-related distance error of the PMD ToF-camera, in: Proc. SPIE, Intelligent Robots and Computer Vision, vol. 6764, doi:10.1117/12.752808, 2007.

[11] D. Falie, V. Buzuloiu, "Distance errors correction for the Time of Flight (ToF) Cameras", Fourth European Conference on Circuits and Systems for Communications (ECCSC'08), (pp. 121-124), Bucharest, Romania , 2008.

[12] D. Falie, V. Buzuloiu, "Further investigations on ToF cameras distance errors and their corrections", Fourth European Conf. on Circuits and Systems for Communications (ECCSC'08), (pp. 125-128), Bucharest, Romania, 2008.

[13] D. Falie, L. David, M. Ichim, Statistical Algorithm for Detection and Screening Sleep Apnea, in: Int. Symp.on Sig., Circ., and Systems (ISSCS), 2009, Iassy, pp.45–48., vol. 1, IEEE Cat. Nr. CFP09816-PRT.